\documentclass[twoside,11pt]{article}
\usepackage[abbrvbib, preprint]{dmlr2e}

\usepackage{lastpage}
\dmlrheading{23}{2023}{1-\pageref{LastPage}}{9/23; Revised 10/23}{11/23}{21-0000}{Andrey Ustyuzhanin, Harald Carlens} % Insert the dates and author names. This is not relevant if it is yet being reviewed, i.e., \documentclass[twoside,11pt, preprint]{article}
\def\openreview{} % Insert correct link to OpenReview for camera-ready version
\ShortHeadings{AI Competitions and Benchmarks: Competition platforms}{Ustyuzhanin, Carlens}
\firstpageno{1}

\usepackage{pifont}% http://ctan.org/pkg/pifont
\newcommand{\cmark}{\ding{51}}%
\begin{document}

\title{AI Competitions and Benchmarks: Competition platforms}

\author{\name Andrey Ustyuzhanin \email andrey.u@gmail.com \\
       \addr Constructor University\\
       Bremen, Germany
       \AND
       \name Harald Carlens \email harald@mlcontests.com \\
       \addr ML Contests
}

% \editor{My editor}

\maketitle

\begin{abstract}%   <- trailing '%' for backward compatibility of .sty file
The ecosystem of artificial intelligence competitions is a diverse and multifaceted landscape, encompassing a variety of platforms that each host numerous competitions annually, alongside a plethora of specialized websites dedicated to singular contests. These platforms adeptly manage the overarching administrative responsibilities inherent in orchestrating competitions, thus affording organizers the liberty to allocate greater attention to other facets of their contests. Notably, these platforms exhibit considerable diversity in their operational functionalities, economic models, and community dynamics. This chapter conducts an extensive review of the foremost services in this realm and elucidates several alternative methodologies that facilitate the independent hosting of such challenges.
\end{abstract}
\begin{keywords}
competition platform, challenge hosting services, comparison
\end{keywords}

\section{Platforms for AI competitions}

The majority of AI competition organisers - around three quarters \footnote{158 of 210 competitions in 2022 took place on the major competition platforms \citep{carlens2023state}. The universe of competitions considered here includes only those with meaningful prize money (over \$1,000) or a conference affiliation. } - use a competition platform to host their competition. 

The choice of platform is driven by various considerations. Before we introduce these, and the role we expect a platform to fulfil, it's helpful to return to a definition of the types of competitions we're considering. We can ground our expectations of a competition platform in the Common Task Framework~\citep{donoho50years} (CTF), which lays out the key ingredients of an AI competition:
\begin{enumerate}
    \item A publicly available training dataset, involving a list of feature measurements and a class label for each observation;
    \item A set of enrolled competitors whose common task is to infer a class prediction rule from the training data;
    \item A scoring referee, to which competitors can submit their prediction rule. The referee runs the prediction rule against a testing dataset which is not made available to competitors. The referee objectively and automatically reports the score achieved by the submitted rule.
\end{enumerate}

While these "ingredients" are somewhat specific to supervised learning competitions, it is not too difficult to see how they would generalise to other types of competitions - such as reinforcement learning style competitions, where data sets are replaced with environments. From these ingredients we can then get a list of responsibilities, to be shared between competition organisers and competition platforms:
\begin{itemize}
    \item \textbf{Design}: framing a problem in a way that is amenable to a CTF-style competition
    \item \textbf{Data}: gathering and cleaning data for training and test datasets
    \item \textbf{Discovery}: notifying potential competitors about the competition
    \item \textbf{Admin}: publishing the competition rules and making training data available
    \item \textbf{Engagement}: enabling competitors to be productive and to collaborate
    \item \textbf{Scoring}: accepting submissions, evaluating them, and updating leaderboards
    \item \textbf{Dissemination}: sharing insights from the winning solutions
\end{itemize}

It is possible for all of these responsibilities to be undertaken by competition organisers, or for them all to be outsourced to a competition platform, but in most cases the responsibilities are shared. The decision of which responsibilities are to be outsourced is of primary importance in the choice of competition platform, as some are better suited to certain responsibilities than others\footnote{For example, CodaLab is able to fulfil most of these, but does not provide support with competition design or data preparation.}. Secondly, the exact requirements for each responsibility will further determine the choice of competition platform \footnote{For example, is the competition aiming to reach a broad audience, or is it targeted at a niche research community who can all be reached through a single mailing list? Is it a straightforward supervised learning problem, or does it incorporate adversarial elements or reinforcement-learning style environments for scoring?}. The remaining components of platform choice come down to budget, familiarity with different platforms, and geographical considerations. 

With these responsibilities in mind, in the next section we lay out more detailed criteria and review the main features of several leading platforms:
\begin{itemize}
    \item AICrowd~\citep{aicrowd},
    \item CodaLab~\citep{codalab_competitions_JMLR} by Universit\'e Paris-Saclay,
    \item DrivenData~\citep{drivendata},
    \item EvalAI~\citep{EvalAI} by CloudCV,
    \item Kaggle~\citep{kaggle} by Alphabet Inc,
    \item Tianchi~\citep{tianchi} by Alibaba,
    \item Zindi~\citep{zindi}.
\end{itemize}

%\adrien{We should cite the platforms, as we have done it in CodaLab's paper \url{https://hal.inria.fr/hal-03629462/document}}

This list is not meant to be comprehensive, and is focused on generalist platforms with active communities as of the end of 2022. We give a separate overview of several non-generalist platforms like \url{grand-challenge.org} or \url{numer.ai} that target specific domains or follow a complementary pattern that doesn't strictly adhere to the CTF setup, as well as some non-English language platforms. 

\section{Platform comparison criteria}
We outline the main characteristics that we use for comparison, which are provided roughly in order of the responsibilities listed above.

\textbf{Design support:} platforms vary in the amount of support they are able to provide to competition organisers in design a competition. Here we are defining the "design" process to cover the initial problem formulation, decisions around the train/test split, and choice of evaluation metrics. This tends to be most important for companies with little in-house data science expertise, and not so relevant for researchers with specific problems in mind. 

\textbf{Data support:} some platforms help competition organisers gather and clean the data they need to run a competition, as well as transforming it into a format that is convenient for competitors to use. 

\textbf{Registered users:} the total number of users registered on a competition platform gives a good indication of the size of the audience that can be reached. This is particularly important for competitions looking to reach a broad audience, or competitors who are not already familiar with the problem area or organiser. 

\textbf{Typical entries:} another way of estimating the community activity on a platform, looking at the number of entries (i.e. number of distinct teams with valid submissions) provides an estimate of how many entries a competition organiser might hope to get on a given platform. This accounts for the fact that some platforms have a more engaged user-base than others, and have better discoverability mechanisms for existing users to find new competitions. 

\textbf{Code sharing:} some platforms allow structured code-sharing through features like Kaggle's Notebooks, or participants can embed their solution as an external Notebook or github repository/commit. This allows other participants to easily reproduce and build on their solutions. Open community collaboration in this way can be a valuable feature for complicated or novel challenges. Such a feature also supports reproducibility.

\textbf{Submission code evaluation:} the most straightforward way to run a competition is to ask participants to submit a set of predictions, and compare those against some "ground truth" values using a loss metric. Some platforms allow for competitions where participants submit code that is then run on the platform side to generate predictions against unseen data. This allows organisers to do things like impose compute budget constraints on submissions, and vet submissions for compliance with the rules. It also changes the nature of the competition, since participants have less knowledge about the distribution of test set examples than they do in the case where they have access to test set features. Most platforms that support code submissions can support it in any language, though support for Python and R tends to be better than for other languages. 

\textbf{Custom metrics:} some platforms or offerings are able to support only "common" metrics like mean squared error or cross-entropy loss. The ability to implement custom metrics is important for many challenges, especially those looking to capture particular trade-offs. Some platforms allow choosing just one among many predefined metrics; some allow for custom implementations. Some platforms charge an additional cost for implementing non-standard metrics. 

\textbf{Staged challenges:} sometimes challenges with datasets taken from a depth of scientific domain might look too mysterious for the community at the beginning, so it helps to split the competition into smaller chunks. Also, it helps to mitigate risks of data leakage by adding a preliminary stage and testing the competition settings. Thus, the platform capable of running a competition in several steps will help people keep the context and smoothly transfer knowledge of the best solutions.

\textbf{Private challenge evaluation:} Data privacy is a sensible industry and even fundamental science issue. Some platforms allow participants' solutions evaluation using an organizer's dedicated machines. Thus with the help of such a feature, one can set up a challenge without sharing restricted datasets even with platform owners.

\textbf{Reinforcement Learning (RL) evaluation:} running participants' RL agents on the platform's side is inherently more complex than running a metric evaluation script across a vector of predictions, and not all platforms support this. Computational cost for these types of competitions is not only often higher than for supervised learning problems, but also more unpredictable - since RL evaluation episode lengths can depend on the success of an agent. Supporting multi-agent environments or tournament-style evaluations are an additional challenge, and we do not evaluate this ability in our analysis. 

\textbf{Human evaluation:} some challenges do not have ground-truth labels in the data. For example, a dialogue bot evaluation requires communication with a living person. Some platforms enable the use of human-evaluation platforms, such as Amazon Mechanical Turk (see below), to check a submission's score. Such an extension is a substantial stretching of the CTF. 

\textbf{Solution publication:} some platforms consistently interview top-ranked participants about their solutions and approaches, and distill these interviews into public blog posts that serve as a lasting resource after the competition ends. Sometimes these are accompanied by detailed technical reports produced by participants, and public code repositories for winning solutions. 

\textbf{Run for free:} most platforms charge a fee for competition hosting. The exact cost usually depends on the range of services offered. Some platforms offer a free "self-service" offering, allowing organisers to set up a completely self-managed competition. 

\textbf{Open-source:} for some platforms, the code that runs them is open-source. In most cases a competition platform fulfils a service, and challenge organisers do not have an interest in changing the platform's functionality. However, being able to access a platform's source code can help organisers assess the pace of development on the platform, verify details of the platform evaluation mechanics, or allow organisers to run their instance on their own premises for a local event with private datasets. It also allows organisers to add features to the platform themselves. 

\section{Platform Comparison}
An overview of platforms as measured by the criteria above is presented in Table~\ref{tab:platforms}. Features which we were able to verify as being supported are marked as \cmark, and where possible these were confirmed with the team running the platform. In some cases where we could not find public documentation of a feature and we did not receive any response from the platform operators, it is possible that we have incorrectly marked features as unavailable. Estimates for the number of users and typical number of entries reflect competition activity in 2022~\citep{carlens2023state}. 

% \dstarr
\begin{table}
%\tbl{Platform overview.}{%
\begin{tabular}{l c @{\hspace{0.7ex}} c @{\hspace{0.7ex}} c @{\hspace{0.7ex}} c @{\hspace{0.7ex}} c @{\hspace{0.7ex}} c @{\hspace{0.7ex}} c @{\hspace{0.7ex}} c }
\hline Criteria                 & AICrowd   & CodaLab   &  DrivenData	&	EvalAI	&	Kaggle  & Tianchi   & Zindi \\
\hline Design support           & \cmark    & -         & \cmark        & -         & \cmark    & -         & \cmark \\  
\hline Data support             & \cmark    & -         & \cmark        & \cmark    & \cmark    & -         & \cmark \\  
\hline Registered users         &   60k 	& 120k 	    & 100k 	        & 30k 	    & 10m       & ?\footnote{We were unable to find an estimate of Tianchi's registered user count. We contacted the platform operators but did not receive a response.}         & 60k \\
\hline Typical entries          &   30 	    & 30 	    & 50 	        & 10 	    & 1,200     &  50       & 180 \\
\hline Code sharing             & \cmark    & \cmark    & -             & \cmark    & \cmark    & \cmark    & - \\
\hline Code evaluation          & \cmark    & \cmark    & \cmark        & \cmark    & \cmark    & \cmark    & - \\
\hline Custom metrics           & \cmark    & \cmark    & \cmark        & \cmark    & \cmark    & \cmark    & \cmark \\
\hline Staged challenge         & \cmark    & \cmark    & -             & \cmark    & -         & -         & - \\
\hline Private evaluation       & -         & \cmark    & -             & \cmark    & -         & -         & - \\
\hline RL-friendly              & \cmark    & \cmark    & -             & \cmark    & \cmark    & -         & \cmark\footnote{Zindi's RL support is not as comprehensive as some other platforms. Zindi has hosted some RL competitions in the past, with policy evaluations taking place on the organiser's servers. } \\  
\hline Human evaluation         & \cmark    & -         & -             & \cmark    & -         & -         & - \\
\hline Solution publication     & -         & -         & \cmark        & -         & -\footnote{While solution write-ups aren't systematically published, Kaggle tends to publish Discussion threads linking to posts describing top solutions.} & ? & \cmark \footnote{Zindi often publishes "Meet the Winners" posts interviewing winners and summarising their approaches: \url{https://zindi.africa/learn/meet-the-winners-of-the-swahili-audio-classification-hackathon}} \\  
\hline Run for free             & -         & \cmark    & -             & \cmark    & \cmark\footnote{Kaggle's free "Community Competitions" are  more limited than their full competitions offering, and do not allow code submissions, custom evaluation metrics, or prize money. } & - & \cmark \footnote{Free competitions can only be run by users enrolled in the Zindi Ambassador program} \\
\hline Open-source              & -         & \cmark\footnote{\url{https://github.com/codalab/codalab-competitions}}        & -     & \cmark\footnote{\url{https://github.com/Cloud-CV/EvalAI}}     & - & - & - \\
\hline Founded         	        &   2017 	& 2013 	    & 2014 	        & 2017      & 2010	    & 2014     & 2018 \\
\hline
\end{tabular}
\caption{Platform overview}
\label{tab:platforms}
\end{table}

% \harald{Todo: review how we're defining the "solution publication" criterion. DrivenData is the only one that does it consistently, but many of the others seem to be able to offer it if it's important to the organiser. Maybe we need a 1-3 star rating for this? And possibly the same for code sharing. }

Here are some highlights of the platforms included in the comparison. 

\textbf{AIcrowd} started as a research project at EPFL, and is now one of the top five competition platforms by total prize money. It has hosted several official NeurIPS competitions as well as many reinforcement learning competitions.

\textbf{CodaLab} is an open-source competitions platform, with an instance maintained by Université Paris-Saclay. Anyone can sign up and host or take part in a competition. Free CPU resources are available for inference, and competition organisers can supplement this with their own hardware. CodaLab is friendly to a variety of challenges: from online data science classes/hackathons to competitions affiliated with leading conferences. CodaLab is suitable to competition organisers who have a clear idea of the competition they want to run, and can be self-sufficient when it comes to technical and marketing aspects. 

\textbf{DrivenData} focuses on running competitions with social impact, and has run competitions for NASA and other organisations. DrivenData stands out for its thorough post-competition reports and permissively licensed solution code publication\footnote{\url{https://github.com/drivendataorg/competition-winners}}. 

\textbf{EvalAI} is built by a team of open source enthusiasts working at CloudCV, a consulting company that aims to make AI research reproducible and easily accessible. With the platform's help, they reduce the entry barrier for research and make it easier for researchers, students, and developers to design and use state-of-the-art algorithms as a service. It is known for running many competitions involving human-in-the-loop evaluations.

\textbf{Kaggle} was acquired by Google in 2017 and has the largest community of all the platforms, having recently reached 10 million users\footnote{\url{https://www.kaggle.com/discussions/general/332147}}. As well as hosting competitions, Kaggle allows users to host datasets, notebooks, and models. Kaggle has a progression system \footnote{\url{https://www.kaggle.com/progression}} which incentivises users to compete, collaborate, share code, and contribute to community discussions. Aside from its main competitions, which can be expensive to run, it is also possible to run a "Community Competition" for free, with limitations around discoverability and evaluation metrics. Costs for running a fully fledged competition with prize money on Kaggle can vary significantly, from a few thousand dollars for the most basic research competitions eligible for Kaggle grants, to well over \$100k for highly customised featured competitions requiring significant development work from Kaggle. 

\textbf{Tianchi} is a competitions platform run by Alibaba, with similar functionality to Kaggle, including running kernels and earning points. Challenges can quickly gain several thousand participants. The primary audience is Chinese, and not all competitions include English documentation.

\textbf{Zindi} is focused on connecting organisations with data scientists in Africa. As well as online competitions, Zindi also runs in-person hackathons and community events.

There are some other platforms which are also worth mentioning:
\textbf{Hugging Face} launched its Competitions\footnote{\url{https://huggingface.co/competitions}} platform in February 2023, alongside its well-established Model Hub and widely-used open source machine learning repositories.
\textbf{bitgrit}\footnote{\url{https://bitgrit.net/competition/}} is an AI competition and recruiting platform founded in 2017, with over 25,000 registered users. 
\textbf{Xeek}\footnote{\url{https://xeek.ai/challenges}} is a competition platform run by Shell's Studio X, which is focused on enabling innovation in the energy sector. Many of the competitions run on this platform so far have been targeted at solving specific business problems, and so the winning solutions are not generally shared publicly. 
\textbf{Analytics Vidhya}~\citep{analyticsvidhya} is a data science community which hosts courses and competitions. It was founded in 2013, and has over 1.5m members, however it is unclear to what extent the platform is now focused on courses and certifications over competitions. The authors contacted Analytics Vidhya's team for more information, but did not get a response.

\section{Non-English language platforms}\label{non_english}
The comparison above is focused on English-language platforms. While the authors are less familiar with platforms in other languages, this section is an attempt at covering platforms in regions where the main common language of their audience is not English. 
As already mentioned, the most notable \textbf{Chinese} competition platform is Tianchi. Other Chinese platforms worth mentioning are: Data Castle\footnote{\url{https://challenge.datacastle.cn/v3/cmptlist.html}}, Kesci\footnote{\url{https://www.kesci.com/}}, Bien Data\footnote{\url{https://www.biendata.xyz/}}, and Data Fountain\footnote{\url{https://www.datafountain.cn/}}. The \textbf{Japanese} platform Signate\footnote{\url{https://signate.jp/}} and the company behind it collaborate with industries, government agencies, and research institutes in various domains to resolve social issues. The \textbf{Russian} community, Open Data Science\footnote{\url{https://ods.ai/}} runs competitions, as well as including organizing events and finding joint projects for researchers, engineers, and developers around Data Science. 
All these platforms above have a reasonably developed community; however, to join those, one needs to be fluent in the corresponding language. 

\section{Domain-specific platforms}\label{domain_specific}
Several platforms host regular challenges on domains in a specific branch of science or industry, or within a more narrow scope than the Common Task Framework. For example, in the finance domain, the \textbf{Numerai}\footnote{\url{https://numer.ai/}} fund draws its strategy from crowd-sourced predictions submitted to regular tournaments. Participants aim to predict stock market movements from obfuscated data. Numerai states that it has paid out over \$30m to its data scientist collaborators. It is worth noting that reward eligibility in Numerai tournaments requires staking Numerai's NMR cryptocurrency token, exposing participants to potential losses, unlike most other platforms listed here.

\textbf{Grand Challenge}\footnote{\url{https://grand-challenge.org/}} is a platform for the end-to-end development of machine learning solutions in biomedical imaging. It has successfully run over a hundred challenges, and allows researchers to host custom algorithms that can be used for performance assessment on new datasets and crowd-sourcing activities called \textit{reader studies}. 

\textbf{NASA Tournament Lab}\footnote{\url{https://www.nasa.gov/coeci/ntl}} (NTL) facilitates the use of crowd-sourcing to tackle NASA challenges. NASA's researchers, scientists, and engineers have launched numerous crowd-sourcing projects through the NTL, seeking novel ideas or solutions to accelerate research and development efforts in support of the NASA mission. The NTL offers a variety of open innovation platforms that engage the crowd-sourcing community to improve solutions for specific, real-world problems being faced by NASA and other Federal Agencies.

\textbf{microprediction}\footnote{\url{https://www.microprediction.com/competitions}} is an open-source competition platform that runs ongoing time-series prediction competitions, allowing anyone to publish "streams" of time-series data with associated rewards, incentivising others to predict upcoming data on those streams.

\textbf{Unearthed}\footnote{\url{https://www.unearthed.solutions}} is a platform that hosts competitions aimed at making the energy and resources industry more efficient and sustainable. Challenges often involve a mixture of domain knowledge and data science skills. 

% \harald{Todo: add reference to codabench/benchmarks chapter once that's been written}

\section{Alternative approaches and adjacent services}

The platforms above are the most notable ones implementing competitions broadly in line with the Common Task Framework~\citep{donoho50years}. However, they are far from the only options for collaborative research. Below is a list of platforms and services that rely on different assumptions and implement interaction protocols that turn out to be suitable for research goals in some scientific domains. 

\textbf{Amazon Mechanical Turk (AMT)\footnote{\url{https://www.mturk.com/}}:} a marketplace for completion of virtual tasks that require human intelligence. Businesses or academic researchers regularly use it to label data that can later be used for training ML algorithms. AMT has been around for more than 15 years. Major companies like Google and Microsoft have similar versions of such marketplaces.

\textbf{Zooniverse\footnote{\url{https://www.zooniverse.org/}}:} While AMT focuses on generic tasks like reading labels from images, captcha translation, listening comprehension, and tagging inappropriate photos, Zooniverse builds a community of people interested in contributing their efforts and intelligence to scientific research advances. It provides participants with unlabelled datasets from various scientific branches: biology, climate, history, physics, etc. Those datasets require human intelligence to label and understand the scientific assumptions of the domain and phenomena presented. Participation in real-science research can motivate people quite significantly. In some cases, discussions between scientists and Zooniverse participants lead to new scientific discoveries~\citep{clery2011galaxy}.

\textbf{OpenML\footnote{\url{https://www.openml.org/}}:} is an online machine learning platform for sharing and organizing data, machine learning algorithms, and experiments. The founders of the platform are passionate about the comparison of different ML methods. Thus they have created a service that allows running an algorithm across several datasets and systematically comparing its performance. While there are no private leaderboards, every check is systematically performed via system API and protocol. Thus new experiments are immediately compared to state of the art without always having to rerun other people’s experiments. The recent development of OpenML involves the design of an AutoML evaluation framework for a broad spectrum of datasets.

\textbf{ML Collective\footnote{\url{https://mlcollective.org/}}}: (MLC) is an independent, nonprofit organization with a mission to make research opportunities accessible and free by supporting open collaboration in machine learning (ML) research. Jason Yosinski and Rosanne Liu founded MLC at Uber AI Labs in 2017 and, in 2020, it moved outside Uber. The group aims to build a culture of open, cross-institutional research collaboration among researchers of diverse and non-traditional backgrounds. Thus, the outcome of the cooperation is the natural growth of participating researchers through discussion and publishing process participation. As of mid-2022, the community is more than 3 thousand ML researchers sharing collaborative research values.

\textbf{PapersWithCode\footnote{\url{https://paperswithcode.com/}}:} organizes access to scientific papers from the leading Machine Learning conferences and links to known implementations of the methods described in such articles. The service also compares different methods of solving several tasks in the form of a leaderboard where entries are linked to particular implementations. The diversity of such leaderboards has grown immensely in the past few years. With the help of this platform, one can find the most current state of the art to the problem of interest and read details of the method in the companion paper.

\textbf{InnoCentive\footnote{\url{https://www.innocentive.com/}}:} is an innovative hub for a new kind of problem-solving. It describes the framework of "Challenge Driven Innovation" (CDI) that helps reformulate a task or opportunity at hand into a series of modules or challenges addressed later by a network of participants. CDI framework is much broader than CTF. Thus Innocentive enjoys various challenges, including Brainstorming, Design, Prototyping, and Algorithm development. The platform has been around for over a decade. It links over half a million solvers and spans dozens of industries. 

\textbf{Seasonal events:} there are many yearly data analysis events organized around the world. Usually, those are hosted by universities and attract quite a significant number of participants. International Data Analysis Olympiad (IDAO)\footnote{\url{https://idao.world}} is just a single example among many others\footnote{Data Mining Cup, \url{https://www.data-mining-cup.com/}},\footnote{ASEAN Data Science Explorers~\url{https://www.aseandse.org/}\label{fn:asean}}. IDAO has engaged several thousand participants across almost a hundred countries each year since 2019. Interesting and unique challenges might fit such events very well. Besides reaching out to a big community, organizers usually run a series of events, including online and offline interactions with the participants. 

\textbf{Hugging Face\footnote{\url{https://huggingface.com/}}:} has quickly become an important resource for the machine learning research community. As well as maintaining several important open-source libraries, most notably its Transformers\footnote{\url{https://github.com/huggingface/transformers}} library with widely-used implementations of NLP models, it also hosts Models and Datasets, enabling easy access to model parameters and training datasets for new research and reproducing others' results.

\textbf{Google Colab\footnote{\url{https://colab.research.google.com/}}:} a hosted notebook solution with support for CPU/GPU/TPU accelerators and sharing via GitHub or Google Drive, Google's Colab service enables interactive code-sharing and eases reproducibility. It significantly lowers the bar for researchers to interact with code or libraries that are not within their domain of expertise, by enabling them to run and edit code without needing to worry about maintaining environments or installing libraries. It can be a useful place for competition organisers to share code examples with potential participants, allowing them a frictionless way to explore a competition. 

\textbf{ML Experiment Tracking Tools:} Tools like MLflow\footnote{\url{https://mlflow.org/}} (open source), W\&B\footnote{\url{https://wandb.ai/}}, Comet\footnote{\url{https://www.comet.com/}}, Neptune\footnote{\url{https://neptune.ai/}} enable distributed research teams to easily share their experimental results within their team or to a public audience. These can serve as a more useful record of experimental results than local tools like TensorBoard or simple text-based logs. 

\textbf{Other:} There are many different venues for interactions between science and citizens. In his book ``Reinventing Discovery: The New Era of Networked Science''~\citep{nielsen2020reinventing}, Michael Nielsen gives a good overview. An interesting example of such interaction is the design of a network of micro-prediction agents that follow a specific question-answering protocol. Authors of those agents get rewards for providing correct answers. Such protocol incentivizes the participants to come up with better algorithms and suitable external data sources~\citep{cotton2019self}. A broader list of citizen-science projects is, of course, available at Wikipedia~\citep{wiki_crowdsourcing}.

\section{Independently hosted competitions}\label{independent}
As we've seen, most competition organisers choose to host their competitions on a competition platform. However, others have shown that it's still possible to "self-host" competitions. Here we give some examples of notable independently hosted competitions. 

\textbf{MIT Battlecode}\footnote{\url{https://battlecode.org}} is an annual competitive real-time strategy game where players need to write code to manage a robot army. The first iteration took place in 2003, predating all currently active competition platforms. Anyone can participate, but only student teams (from any university) are eligible for prizes. Recent sponsors include game studios and quantitative trading firms. MIT students participating in Battlecode are eligible for credits, as it is a registered course. 

\textbf{Real Robot Challenge}~\citep{pmlr-v176-bauer22a}\footnote{\url{https://real-robot-challenge.com}} is a competition involving dexterous manipulation tasks using robot hands. Evaluation takes place on physical robots. Participants are provided with software simulation environments to train their policies, and are able to submit their policies for physical evaluation. The organising team had to do significant development work in order to be able to accept submissions to run on their physical robots, and they decided to self-host the whole competition since the additional work to build their own leaderboard was deemed easier than integrating with an existing competition platform. 

\textbf{The Humanoid Robot Wrestling Competitions}\footnote{\url{https://webots.cloud/competition}} are a series of simulated robotics competitions in the webots simulator. The organisers built their own leaderboard management framework on top of GitHub Actions, enabling anyone with a GitHub account to take part in the competition. Evaluations are run automatically on a dedicated server owned by Cyberbotics Ltd, the company behind the simulator, whenever a competitor pushes a code change to their GitHub repository. Participants' code can stay private; participants just need to add the competition organiser's GitHub account as a collaborator on their repository. The organisers helpfully shared their competition template\footnote{\url{https://github.com/cyberbotics/competition-template}} under a generous open-source licence, enabling others to run competitions like this with minimal additional setup.

\section{Choosing the best platform for a competition}

Given the set of platforms available, choosing the one best suited to a particular competition is not trivial. We hope that table ~\ref{tab:platforms} can be a helpful resource for competition organisers. 
In addition to this, we can provide some general advice.

For companies with limited in-house data science expertise or tech resources, it makes sense to choose a platform which offers support with challenge design and data preparation. While these platforms can require a larger budget than alternative options, they are often able to leverage their existing significant user-base to engage desirable and capable competitors, resulting in more and higher-quality submissions than might otherwise be possible. This reduces the pressure on competition organisers to promote the competition themselves. 

Competition organisers with a limited budget will generally have to take on the challenge design and data preparation work themselves. In these cases, unless an additional competition sponsor can be found, using a platform with free competition hosting options will likely be desirable. In order to aid with discoverability on the free hosting options - helping potential participants find the competition - organisers might want to try to get their competition mentioned in relevant newsletters, or submit their competition to a competition listing site like ML Contests\footnote{The website \url{https://mlcontests.com}, maintained by one of the authors of this article, hosts a list of currently ongoing competitions with meaningful prize money or conference affiliations. It helps organisers of smaller competitions find competitors, and helps competitors discover competitions across all platforms, and is open for submissions from anyone.} if they are trying to reach a broad audience. 

Teams organising competitions with particular requirements - reinforcement learning environments, data privacy restrictions, or human-in-the-loop evaluation - are more restricted in their choice of platforms than "vanilla" supervised learning competitions. It's worth noting that even if platforms don't officially list certain features, sometimes they are able to accommodate additional requirements - so it can be worth having an exploratory conversation before ruling platforms out, as long as sufficient budget is available to compensate platforms for any additional development they might need to do. 

Competitions targeted at niche communities might benefit from the relevant exposure they would get on a domain-specific platform  (see section \ref{domain_specific}). Similarly, competitions targeting participants with certain language skills or located in particular geographic areas might take this into account when choosing a platform (see section \ref{non_english}). 

Only organisers of the most idiosyncratic competitions or those with significant in-house resources would likely find it preferable to run a competition without making use of any platform. We mention some examples of these in section \ref{independent}.

\section{Conclusion}
This chapter presents an overview of the most popular AI competition platforms. It gives a summary of each of the platforms, introduces key criteria for platform comparison, and uses these to provide a simple comparison table that we hope will be a useful reference for any competition organiser looking to find the most suitable service for running their competition and maximising its potential impact.  

%%%%%%%%%%%%%%%%%%%%%%%%%%%%%%%

%Here is a citation ~\citep{chow:68}.

% Acknowledgements and Disclosure of Funding should go at the end, before appendices and references

%All acknowledgements go at the end of the paper before appendices and references.
%Moreover, you are required to declare funding (financial activities supporting the submitted work) and competing interests (related financial activities outside the submitted work).
% More information about this disclosure can be found on the DMLR website.

\acks{
The work presented in this book chapter was undertaken as a community collaboration and did not receive any external funding.
}


% Authors must include a Broader Impact Statement, which should provide a concise, tangible portrayal of both the potential positive and negative societal consequences of their work. We refer to the submission guidelines for further details.

% \impact{
% This book offers educational benefits, emphasizing the pedagogic value of challenges in AI. The content fosters community engagement, showcasing how challenges can drive collective innovation. Moreover, it reinforces the bridge between academia and industry, highlighting the transformative role of challenges in transitioning research to real-world applications. The book encourages the community to design competitions and benchmarks well aligned with real-world needs and ethical codes.
% }

%%% References %%%
\vskip 0.2in
\bibliography{main}
\newpage
%%%%%%%%%%%%%%%%%%

%\appendix
%\section{}
%\label{app:theorem}

% Note: in this sample, the section number is hard-coded in. Following
% proper LaTeX conventions, it should properly be coded as a reference:

%In this appendix we prove the following theorem from
%Section~\ref{sec:textree-generalization}:

%In this appendix we prove the following theorem from
%Section~6.2:

%\noindent
%{\bf Theorem} {\it Let $u,v,w$ be discrete variables such that $v, w$ do
%not co-occur with $u$ (i.e., $u\neq0\;\Rightarrow \;v=w=0$ in a given
%dataset $\dataset$). Let $N_{v0},N_{w0}$ be the number of data points for
%which $v=0, w=0$ respectively, and let $I_{uv},I_{uw}$ be the
%respective empirical mutual information values based on the sample
%$\dataset$. Then
%\[
%	N_{v0} \;>\; N_{w0}\;\;\Rightarrow\;\;I_{uv} \;\leq\;I_{uw}
%\]
%with equality only if $u$ is identically 0.} \hfill\BlackBox

%\section{}

%\noindent
%{\bf Proof}. We use the notation:
%\[
%P_v(i) \;=\;\frac{N_v^i}{N},\;\;\;i \neq 0;\;\;\;
%P_{v0}\;\equiv\;P_v(0)\; = \;1 - \sum_{i\neq 0}P_v(i).
%\]
%These values represent the (empirical) probabilities of $v$
%taking value $i\neq 0$ and 0 respectively.  Entropies will be denoted
%by $H$. We aim to show that $\fracpartial{I_{uv}}{P_{v0}} < 0$....\\

\end{document}